\documentclass[11pt]{article}
\usepackage{coling2020}
\usepackage{times}
\usepackage{url}
\usepackage{latexsym}

\usepackage{amsmath}
\usepackage{graphicx} 
\usepackage{booktabs}
\usepackage{tikz-dependency}
\usepackage{tikz}

\usepackage{comment}

\colingfinalcopy 

\title{Beyond The Text: Analysis of  Privacy Statements  through  Syntactic and Semantic Role Labeling}

\author{Yan Shvartzshnaider \\
  New York University \\
  {\tt yansh@nyu.edu} \\\And
  Ananth Balashankar\\
 New York University \\
 {\tt ananth@nyu.edu}\\\And
 Vikas Patidar\\
 New York University\\
 {\tt vp1274@nyu.edu}\\\AND
 Thomas Wies\\
 New York University \\
 {\tt wies@cs.nyu.edu}\\\And
 Lakshminarayanan Subramanian\\
 New York University\\
 {\tt lakshmi@cs.nyu.edu}
 \\}

\date{}

\begin{document}
\maketitle
\begin{abstract}
This paper formulates a new task of extracting privacy parameters from a privacy policy, through the lens of Contextual Integrity, an established social theory framework for reasoning about privacy norms. Privacy policies, written by lawyers, are lengthy and often comprise incomplete and vague statements. In this paper, we show that traditional NLP tasks, including the recently proposed Question-Answering based solutions, are insufficient to address the privacy parameter extraction problem and provide poor precision and recall. We describe 4 different types of conventional methods that can be partially adapted to address the parameter extraction task with varying degrees of success: Hidden Markov Models, BERT fine-tuned models, Dependency Type Parsing (DP) and Semantic Role Labeling (SRL). Based on a detailed evaluation across 36 real-world privacy policies of major enterprises, we demonstrate that a solution combining syntactic DP coupled with type-specific SRL tasks provides the highest accuracy for retrieving contextual privacy parameters from privacy statements. We also observe that incorporating domain-specific knowledge is critical to achieving  high precision and recall, thus inspiring new NLP research to address this important problem in the privacy domain.
\end{abstract}

\section{Introduction}

A privacy policy informs users about a company's information handling practices. However, privacy policies are lengthy documents, full of incomplete and vague statements that impose a significant cognitive burden on the reader to infer whether a given service respects their privacy~\cite{bhatia2016theory,bhatia2018semantic,reidenberg2015disagreeable}.

This challenge has inspired many recent works in applying natural language processing and machine learning techniques to automatically process privacy policies and retrieve the relevant information~\cite{harkous2018polisis,ravichander-etal-2019-question}. While these efforts help in identifying paragraphs in the privacy policy that mention sensitive information~\cite{evans2017evaluation,bhatia2015towards}, opt-out clauses~\cite{sathyendra2016automatic} or description of data collection practice~\cite{sadeh2014towards}, they focus on the policy as a whole rather than the individual privacy statements that it contains. In particular, they do not aim to identify relevant and often missing contextual information that are critical for unambiguously understanding the scope of individual statements. This paper focuses on a new NLP task that aids the analysis of privacy policies at this more fine-grained level.

To illustrate the problem, consider a typical example of an ambiguous privacy statement: ``{\em {\bf Yahoo} collects {\bf information about your transactions with us and with some of our business partners}, including {\bf information about your use of financial products and services that we offer.}}" At first glance, the statement may seem to provide all the relevant information about a first-party collection of transactional data. However, it in fact misses some crucial contextual information. To understand what is missing, we use the contextual integrity (CI) framework~\cite{nissenbaum2009privacy}.  CI defines privacy as an appropriate flow of information which is expressed in terms of 5 essential \emph{CI parameters}: Sender, Recipient, Subject, Information Type, and Transmission Principle. The latter is a constraint on the information flow expressing the condition under which information is being transferred.
The above statement specifies only 3 out of the 5 necessary parameters (highlighted in bold) -- Subject, Recipient and Information Type. This leaves the sender of the information and transmission principle to the reader's interpretation. In some cases, the relevant missing information appears in different places in the policy, for example, under different sections such as ``When do we collect your information" or ``Our partners". These, however, do not help in contextually positioning the above statement so that the reader can determine whether their expectations have been met. 
 
In this paper, we formulate the new NLP task of extracting the CI parameters from privacy statements (\S~\ref{sec:Task}). 
We describe four different types of conventional methods that have been partially adapted to address this task: Hidden Markov Models, BERT models, Dependency-Type Parsing and CI specific Semantic Role Labeling (\S~\ref{sec:Methods}). Our evaluation of 36 real-world privacy policies shows that a solution combining syntactic dependency type parsing (DP) coupled with type-specific Semantic Role Labeling (SRL) tasks provides the highest accuracy for retrieving contextual privacy parameters from privacy statements (\S~\ref{sec:Evaluation}). We also observe that incorporating domain-specific knowledge is critical and doing so, we successfully extract the relevant CI parameters with F1  score of 80\% or higher.

%

\section{Related Work} 
Several recent efforts have focused on identifying important and relevant privacy statements using constituency parsing ~\cite{sathyendra2017identifying,sathyendra2016automatic,evans2017evaluation}, logistic regression ~\cite{ammar2012automatic} and crowdsourcing~\cite{Wilson2016} techniques. \newcite{harkous2018polisis} trained a machine learning model for querying privacy policies to retrieve relevant passages of information. Specifically, it supports free form questions about data handling practices described in the text and returns the paragraph mentioning the relevant practice. As we discuss in Section~\ref{sec:Task}, our work explicitly looks to map the privacy statement to a fixed set of parameters. We also show that Question Answering (QA) models do not perform satisfactorily when applied to our task.

Similar limitations of the reading comprehension models were observed by \newcite{ravichander-etal-2019-question}, who composed the PRIVACYQA dataset, an annotated corpus consisting of 1750 questions about the contents of privacy policies such as ``What data does this game collect?" and ``Will my data be sold to advertisers?". Our work is inspired by these efforts to provide a dataset of CI parameter annotations and a machine learning model for automatic CI parameter extraction.

In prior work on automatic privacy statement analysis, \newcite{bhatia2016mining} extracted privacy statements on information handling practice such as ``collecting your e-mail address” or “sharing your location"
using typed dependency parser and crowdworker annotations. More relevant to our efforts, \newcite{bhatia2018semantic} applied Semantic Roles theory to manually annotate 5 privacy statements and identify action verbs (action data) such as ``collection", ``retain", ``use", ``transfer" and associated semantic roles that capture who performs the action, how the action is carried out, etc. 

\newcite{shvartzshnaider2019going} crowdsourced  privacy policies annotation to  compare policy versions, identifying missing contextual information and overloading of parameters that contribute to users' inability to understand the prescribed information practices. Our work automates the task of annotating privacy policies with the CI parameters. 

Many other multidisciplinary efforts draw on CI, as the underpinning privacy theory and can benefit from our newly formulated annotation task. Legal scholars and social scientists have used CI to examine existing data sharing practices in companies like Facebook~\cite{hull2011contextual} and Google~\cite{zimmer2008privacy} in order to identify important contextual elements behind users' privacy expectations~\cite{apthorpe2018discovering,martin2016measuring}. In computer science, researchers have used CI to build privacy compliance and verification tools~\cite{barth2006privacy,chowdhury2013privacy}.

\section{CI Parameters Extraction Task}\label{sec:Task}
In this section we formulate the task of extracting relevant CI parameters from privacy policy statements. 

Let us first motivate this task by discussing its applications. To perform an analysis of privacy implications of a given information flow, the theory of CI requires identifying 5 essential parameters: actors (sender, receiver, subject), the type of information (attribute), and condition of the information exchange (transmission principle).  This analysis can help in identifying potentially confusing or misleading statements, e.g., when one of the five parameters such as transmission principle or receiver is missing or ambiguous~\cite{shvartzshnaider2019going}. Furthermore, one can use the identified parameters to formalize the expressed informational norms and privacy rules in formal logic~\cite{shvartzshnaider2019vaccine,datta2011understanding}. These formalisms can in turn be used to build systems that enforce the specified rules or automatically audit information flows to detect rule violations.

The CI parameter extraction task is as follows. Given a privacy statement $stmt$, apply a mapping function $M$ to extract the CI parameters: sender, receiver, subject, attribute, transmission principle:
\begin{align*}
    M(stmt)  = (s,r, sub, att, tp)
\end{align*}
The main challenge behind the task is in identifying the lexical items in the statement that correspond to the contextually relevant values to help downstream NLP tasks perform the privacy analysis. This is not a trivial task as privacy policies are not written with CI in mind. Often, they are written by legal and policy teams whose primary concern is not readability. Many privacy statements are missing essential CI parameters and often comprise syntactically complex sentences~\cite{bhatia2018semantic}. In the absence of an automatic way to extract CI parameters, researchers have employed crowd-sourcing and manual annotation to perform the analysis~\cite{shvartzshnaider2019going}. The results, while promising, are not yet satisfactory and have many challenges.  We provide a motivating example to demonstrate the challenges involved in this task. Consider the privacy statement:
\medskip\\
 \scalebox{0.9}{
    \begin{dependency}[edge style={red,densely dotted}]
        \begin{deptext}
        We \& transfer \& information \& about you \& if \& Yahoo is acquired by or merged with another \& company.\\
        \end{deptext}
        
        \deproot[edge unit distance=1ex]{1}{Sender}
        \deproot[edge unit distance=1ex]{3}{Attribute}
        \deproot[edge unit distance=1ex]{4}{Subject}
        \depedge[edge unit distance=1ex]{5}{7}{TP}
        \wordgroup{1}{5}{7}{to}
        \end{dependency}
        }
\medskip\\
Viewed through the lens of CI, we are interested in answering the following questions: {\em ``Who is transferring?"}, {\em ``What is being transferred?"}, {\em ``Who is the subject?''}, {\em ``Who is the receiver/recipient?"}, {\em ``Why, When and How is the transfer facilitated?"}. The relevant CI parameters are marked in the statement mentioned above. We tried applying an open domain QA model to answer these questions. Table~\ref{tab:QA_F1} shows results of our expeditionary experiment. The overall F1 scores for the QA model indicates poor results for extraction of all CI parameters. Note in our experiment, QA outputs multiple phrase predictions for each of the parameters. For precision, we calculate true positives as a fraction of all positives predicted for each parameter. For recall, we calculate the fraction of true positives to all correct parameters.  

\begin{table}[!ht]
\centering
\begin{tabular}{lccc}
        &  {\bf Recall} &  {\bf Precision} &   {\bf F1} \\
\midrule
    Attribute &    0.21 &       0.14 &  0.17 \\
    Receiver &    0.07 &       0.06 &  0.06 \\
    Sender &    0.03 &       0.02 &  0.03 \\
    Subject &    0.06 &       0.02 &  0.03 \\
    TP &    0.21 &       0.16 &  0.18 \\
\end{tabular}
\caption{Precision, Recall and overall F1 score for QA Comprehension model used for the CI parameter extraction task. The recall and precision values for a parameter are calculated by macro averaging over privacy statements. }
\label{tab:QA_F1}
\end{table}

This result aligns with previous uses of QA in the privacy domain. 
\newcite{ravichander-etal-2019-question} observed that, compared to a human annotator, Question Answering for Privacy Policies using standard reading comprehension models returns relatively poor results in answering specific questions such as ``will my data be sold to advertisers?" and ``what data does this [service] collect?".

These experiences suggest that QA models require additional heuristics to filter the many false positives as a result of them operating on a paragraph level and not on sentence level statements. Thus, we have established that extracting CI parameters using existing off-the-shelf models without significant re-mapping leads to low precision and recall. 

In the next section, we discuss how we can re-purpose existing tasks by leveraging domain expertise in CI to extract these parameters. We further demonstrate that, with a comparable dataset, training an end-to-end supervised learning model does not provide accurate results.  
\begin{figure*}
 \includegraphics[scale=0.35]{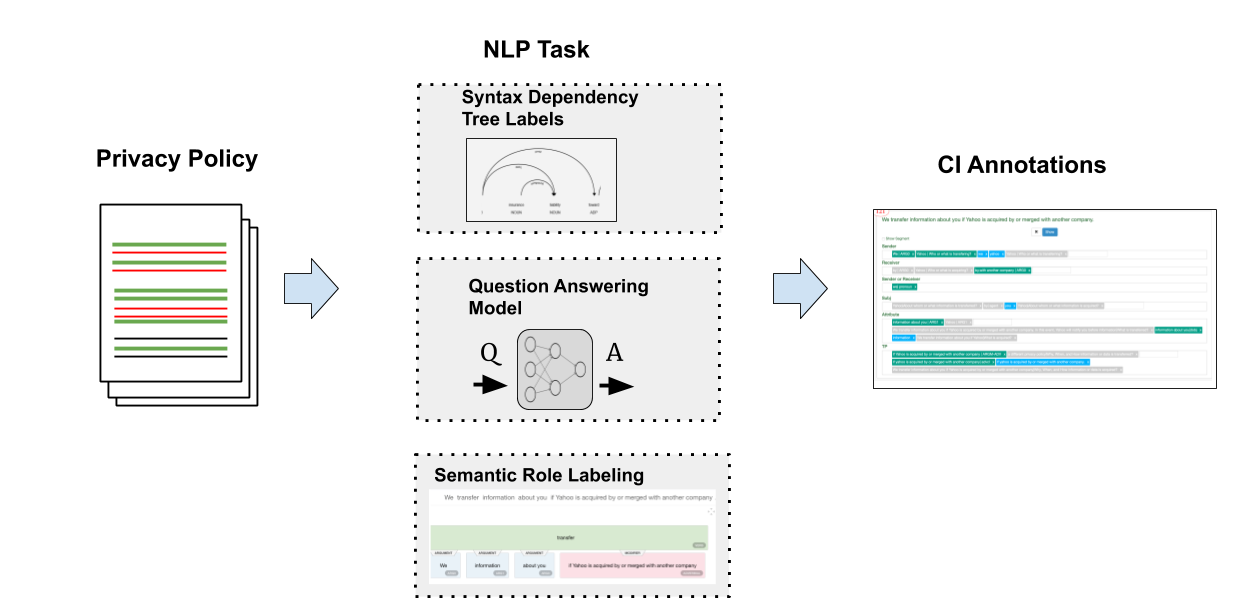}
    \caption{CI Parameters annotation task  pipeline}
    \label{fig:pipeline}
\end{figure*}
\section{Methods}
\label{sec:Methods}

In this section we describe the NLP methods we applied to the CI parameter extraction task:  Hidden Markov Model, BERT, Dependency Parser (DP) and Semantic Role Labeling (SRL). We illustrate the post-processing and the modifications required in off-the-shelf end-to-end neural models to extract CI parameters from privacy policies. Specifically, we focus on Syntactic DP and SRL-based approaches.

\subsection{Hidden Markov Model}

We formulate the CI parameter extraction as a part-of-speech (POS) tagging task  and use a Hidden Markov Model (HMM) probabilistic model~\cite{jurafsky2014speech} for annotating words in a sentence. Specifically, we train a trigram HMM by converting the dataset to CoNLL-2003 format~\cite{sang2003introduction} with CI parameters as the target labels. In our setup, we use 80/20 train-test split, with a training set comprising of 2504 privacy statements and 18533 tokens and a validation set consisting of 626 privacy statements and 5130 tokens. By default, HMM relies on the Markov assumption that the probability of a particular state only depends on the preceding state. However, in order to enrich our HMM model, we  consider the two previous states when predicting the current CI parameter, turning it into a trigram model. Further, we obtain the final transition probability distribution by linearly combining unigram, bigram and trigram probability distributions:
    \begin{align*}
    P({t}_i\mid {t}_{i-1},{t}_{i-2})= \lambda_1 P({t}_i\mid {t}_{i-1},{t}_{i-2}) +  \lambda_2P({t}_i\mid {t}_{i-1}) +  ( 1- \lambda_1 - \lambda_2) P({t}_i)
    \end{align*}
 The parameters $\lambda_1$ and $\lambda_2$ are fine-tuned on the validation set with values 0.42 and 0.48 providing the best results. The Viterbi algorithm~\cite{forney1973viterbi} is used in the decoding phase for the extended model.

\subsection{Bidirectional Encoder Representations from Transformers (BERT)}

We frame the CI parameter extraction task as a sequence-to-sequence transformation problem to fine-tune an advanced BERT model~\cite{DBLP:journals/corr/abs-1810-04805} on our dataset to map a sequence of words in privacy statements to a corresponding sequence of CI tags. 
For training and testing, we transformed  our dataset into the CoNLL2003 format and used  AllenNLP~\cite{gardner2018allennlp} with the train-test split ratio as 80/20 and values of hyperparameters taken from~\cite{gardner2017deep}.

\subsection{Dependency parsing}
Dependency parsing is the task of identifying syntactic roles or dependency types for each of the words in a sentence.  
This involves parsing a sentence and identifying the syntactic structure denoting the grammatical rules that governs a language. Not all the dependency types identified for the English language are relevant in our study.

We use the DP outputs to identify the relevant CI parameters in the privacy statement. To identify CI parameters at a single sentence level using local relationships, we run a typed dependency parser (DP) on the text of the policies. We accept paragraphs as input, split them into sentences and parse each sentence using the Spacy I/O\footnote{https://spacy.io} dependency parser. 
The library~\cite{spacy2} achieves near state-of-the-art performance on most NLP tasks\footnote{https://spacy.io/usage/facts-figures}. We then map the dependency types to specific CI parameters as shown in  Table~\ref{tab:dep_ci}. 
\begin{table}[!h]
\begin{center}
\begin{tabular}{lp{12em}}
\textbf{CI Parameter Type} & \textbf{Dependency types}\\
\toprule
Attribute & {\em dobj}, {\em parataxis}, {\em nsubjpass}\\
Sender/Receiver & {\em nsubj}, pronouns\\
TP & {\em xcomp}, {\em ccomp}, {\em advcl}, {\em oprd}\\
Subject & {\em poss}, {\em agent}\\
\end{tabular}
\end{center}
\caption{Mapping of dependency types corresponding to CI parameters. To represent dependencies we use the Stanford Typed Dependency Manual~\cite{de2011stanford} notations.}
\label{tab:dep_ci}
\end{table}%

For example, for the following statement from the Google privacy policy, the DP praser will return the following dependency type tags (white nodes), which are mapped to corresponding CI parameter (gray nodes):
\medskip\\
 \noindent
 \scalebox{0.9}{
    \begin{dependency}[edge style={red,densely dotted}]
        \begin{deptext}
        When \& you use Google \& services, \&   we \& may collect and process\& information \& about \& your \& actual \& location.\\
        \end{deptext}
        
        \depedge[edge unit distance=1ex]{1}{3}{advcl}
        \depedge[edge unit distance=1.2ex, edge below, label style={fill=gray!40,font=\bfseries,text=black}]{1}{3}{TP}
        \deproot[edge unit distance=1ex]{4}{pron}
        \deproot[edge unit distance=1.2ex, edge below, label style={fill=gray!40,font=\bfseries,text=black}]{4}{Receiver}
        \depedge[edge unit distance=1.2ex]{6}{10}{dobj}
        \depedge[edge unit distance=1.2ex, edge below, label style={fill=gray!40,font=\bfseries,text=black}]{6}{10}{Info. type}
        \deproot[edge unit distance=0.8ex]{8}{poss}
        \deproot[edge unit distance=0.8ex, edge below, label style={fill=gray!40,font=\bfseries,text=black}]{8}{Subject}
        \end{dependency}        
       }
\smallskip\\
\begin{figure*}[t]
\begin{minipage}{.5\textwidth}
    \centering
		\includegraphics[width=0.9\linewidth]{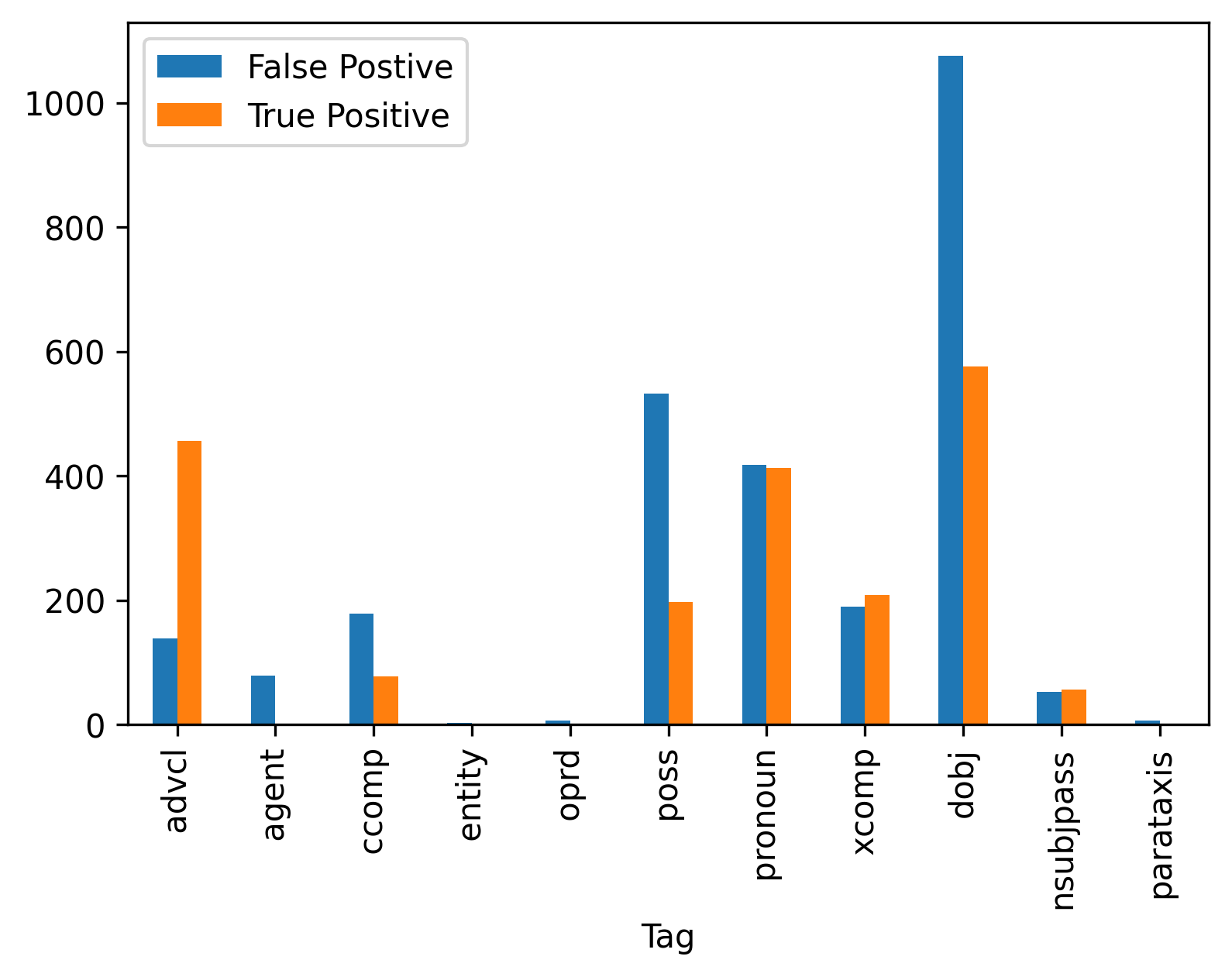}
    \caption{Distribution of True Positives and False \\Positives for each SRL tag}
    \label{fig:DP_tag_dist}
\end{minipage}%
\begin{minipage}{.5\textwidth}
  \centering
	\includegraphics[width=0.9\linewidth]{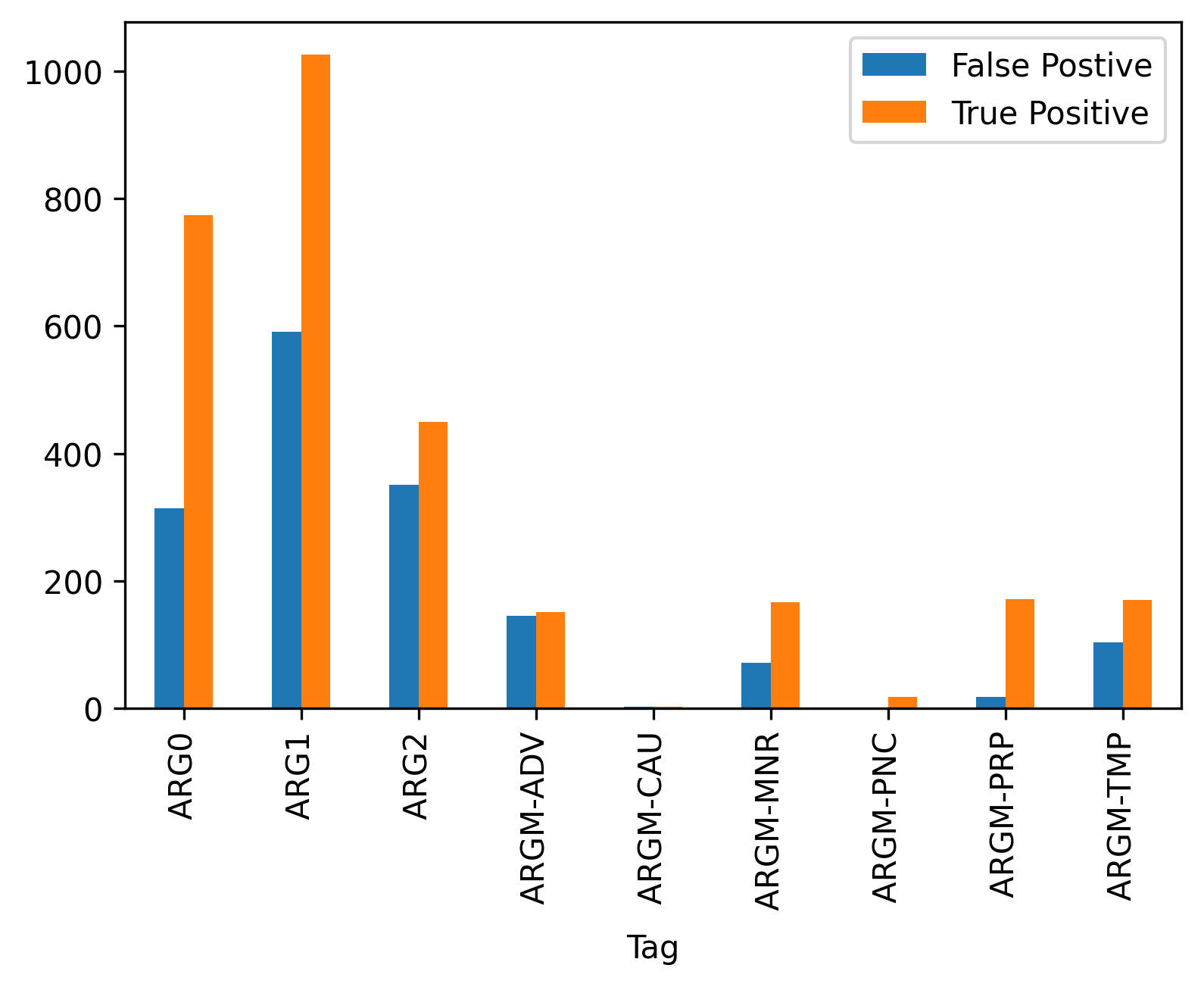}
    \caption{Distribution of True Positives and False \\Positives for each DP tag}
    \label{fig:SRL_tag_dist}
\end{minipage}
\end{figure*}
Note that, as is evident in Table~\ref{tab:dep_ci}, the dependency types cannot distinguish between the parameter of sender and receiver. For this, we defer to the task of SRL to identify based on the semantic meaning of the word. Figure~\ref{fig:DP_tag_dist} shows the percentage of DP tags that are correctly and incorrectly mapped to the CI parameters. This indicates the diversity and coverage of the many tags that map to each of the CI parameters. It also illustrates that the task of extracting CI parameters is not equivalent to that of DP and new conditional information is required to modify DP and solve the task.

\subsection{Semantic Role Labeling}

Semantic Role Labeling is the task of mapping words or phrases in a sentence to a semantic role such as that of an agent, goal, or result~\cite{jurafsky2014speech}. Often, in the classic natural language processing pipeline, this task is considered to have subsumed syntactic and parts-of-speech tasks within it~\cite{tenney-etal-2019-bert}. For example, the task of distinguishing between a sender and receiver can be done through SRL, but not through syntactic DP.

Similar to DP, we map the semantic roles to the relevant CI parameters. Table~\ref{tab:SRL_ci} shows the CI parameter mapping based on a verb's syntactic arguments. For example the verb ``collect" has the following associated arguments (see PropBank corpus~\cite{martha2005proposition}): \verb+ARG0+: agent, entity acquiring something, \verb+ARG1+: thing acquired, \verb+ARG2+: source, \verb+ARG3+: more specific attribute of \verb+ARG1+ being collected, \verb+ARG4+: benefactive.
\smallskip\\
To recover the predicate argument structure of a sentence we use an AllenNLP implementation of the Bidirectional LSTM model~\cite{He2017DeepSR}. 
For example, for the following statement the SRL model returns:  
\medskip\\
 \noindent
  \scalebox{0.9}{
    \begin{dependency}[edge style={red,densely dotted}]
        \begin{deptext}
        We \& {\bf collect} \& technical \& information \&[1em] when \& you {\bf visit} our \& websites\\
        \end{deptext}

        \deproot[edge unit distance=1ex]{1}{ARG0}
        \deproot[edge unit distance=1ex]{2}{V}
        \depedge[edge unit distance=3ex]{3}{4}{ARG1}
        \depedge[edge unit distance=1.5ex]{5}{7}{ARGM-TMP}
         \wordgroup{1}{5}{7}{argm-tmp}
        \wordgroup{1}{3}{4}{arg1}
        \end{dependency}
        \newline
       \begin{dependency}[edge style={red,densely dotted}]
        \begin{deptext}
          or \& use our mobile applications or \&services\\
        \end{deptext}
        \depedge[edge unit distance=1ex]{1}{3}{ARGM-TMP}
        \wordgroup{1}{1}{3}{argm-tmp}

        \end{dependency}   
        }
We then map the arguments onto the CI parameters. In the above example, \verb+ARG0+ is mapped to Recipient. \verb+ARG1+ is an Attribute, and ARGM-TMP is the TP. For each of the verbs these mapping are slightly different, as shown in Table~\ref{tab:SRL_ci}. This mapping, although crude, covers a significant class of privacy policy statements which describe norms of information flows. 

\begin{table}[!ht]
\begin{center}
\scalebox{1}{
\begin{tabular}{lp{7em}p{6em}}
& {\bf ``Sending" action verbs} & {\bf ``Receiving" action verbs}  \\
\midrule
Sender & {\em ARG2 } &  {\em ARG0}\\
Receiver & {\em ARG0} &  {\em ARG2}\\
  &\\
\toprule
Attribute & {\em ARG1}, {\em C-ARG1}\\
TP & \multicolumn{2}{l}{{\em ARGM-TMP}, {\em ARGM-ADV}, {\em ARGM-MNR}}\\ 
   &\multicolumn{2}{l}{ {\em ARGM-PNC}, {\em ARGM-CAU}}
   \\ 

\end{tabular}
}
\end{center}
\caption{Mapping semantic roles (notations) to specific CI parameters.}
\label{tab:SRL_ci}
\end{table}%

\paragraph{CI-related Semantic Frames} The SRL model returns verb-argument predicates for all the identified verbs in a sentence. Some of these verbs are not relevant to information exchange. For example, in the above statement, the verb ``visits" does not convey semantically meaningful information regarding the exchange of technical information. 

To reduce the number of false positives, we provide a list of verbs to the algorithm which highly correlate with information exchanges. It is helpful to think of this approach through the lens of the linguistic theory of Frame semantics~\cite{fillmore1976frame}, which posits that specific meaning of words (frame elements) can be understood only  as part of a particular context (semantic frames). In our approach, we would invoke CI-related semantic frames. 
Specifically, we look for SRL-predicates that are associated with any transfer of information (actual or perceived).  This includes a list of verbs such as ``sending", ``sharing", ``transmitting" and others. 
In addition to invoking a general semantic frame, we differentiate between different roles of associated argument with each predicate. In particular, for predicates like ``sending", ``sharing", ``transmitting" the \verb+ARG2+ is typically associated with the agent role of a ``sender", the \verb+ARG1+ captures what was ``sent" and  \verb+ARG0+ is associated with the receiving agent role. For verbs like ``gather'', ``collect", ``receive", ``acquire"  the roles are reversed: \verb+ARG0+ is typically associated with a ``sending" agent role, the \verb+ARG1+ describes what is ``Received", and \verb+ARG2+  is associated with the ``receiving" agent role. Grouping the verbs signifying a ``sending" or a ``receiving" action helps us map the corresponding arguments to the relevant CI parameters for Senders and Receivers. The mapping for TP and Attribute remains the same for all verbs. Finally, our SRL mapping does not include a semantic role mapping of the Subject parameter. We operate on the assumption that the subject in most statements is the user.

\subsubsection{Clues from CI to Improve SRL} \label{sec: heuristic}

Identifying the arguments for all verbs in the privacy statement results in high recall numbers. Nevertheless, the precision suffers because not all of the verbs need to be invoked. To reduce the number of false positive mappings, we implement an algorithm which analyzes all the relevant SRL verbs to check whether any of them appear as part of the Transmission Principle (TP) relative to another verb. 

For example, in the following statement the SRL model will pick up two predicates (verbs) and corresponding arguments:\smallskip\\
\scalebox{0.9}{
\begin{minipage}{.5\textwidth}
    \centering
	\begin{dependency}[edge style={red,densely dotted}]
        \begin{deptext}
        {\em collect:} \& We \& {\bf collect} \& your \& personal \& information  \\
        \end{deptext}
        \deproot[edge unit distance=1ex]{2}{ARG0}
        \deproot[edge unit distance=1ex]{3}{V}
        \depedge[edge unit distance=1ex]{4}{6}{ARG1}
        \end{dependency}
       \begin{dependency}[edge style={red,densely dotted}]
        \begin{deptext}
              \&[3em]  when \& you are sharing  your\& post.\\
        \end{deptext}
        \depedge[edge unit distance=1ex]{2}{3}{ARGM-TMP}
        \end{dependency}
\end{minipage}%
\begin{minipage}{.5\textwidth}
\vspace{2em}
  \begin{dependency}[edge style={red,densely dotted}]
        \begin{deptext}
        {\em sharing:} \& We collect \& your personal \& information  \\
        \end{deptext}
        \end{dependency}
        \newline
       \begin{dependency}[edge style={red,densely dotted}]
        \begin{deptext}
              \&[3em] when  \&[2em] you \& are \&  {\bf sharing} \& your\& post.\\
        \end{deptext}
        \deproot[edge unit distance=1ex]{2}{ARGM-TMP}
        \deproot[edge unit distance=1ex]{3}{ARG0}
        \deproot[edge unit distance=1ex]{5}{V}
        \depedge[edge unit distance=2.5ex]{6}{7}{ARG1}
        \end{dependency}  
  
\end{minipage}
}
\smallskip\\
 These arguments will be mapped to CI parameters, as described in the previous section. The verb ``share" is redundant in this context since it is part of the TP of the verb ``collect." Once we identify the \textit{redundant} verb, we ignore all arguments associated with it, i.e., our algorithm does not consider these results. We do keep those parameters that overlap with the parameters produced by non-redundant verbs. For instance, in our example we ignore the verb ``share" and the associated with it arguments. Specifically, the [{\bf ARG0:} you] and the [{\bf ARG1:} your post] which otherwise will be mapped to a CI sender and attribute parameters, respectively.

\section{Evaluation}
\label{sec:Evaluation}

We perform automatic annotation of 36 policies of the OPP-115 Corpus~\cite{wilson2016creation}. The corpus' privacy policies were annotated to specify data practices mentioned in each of the segments of the policy.  We limit our CI parameter extraction to labeled  segments of the policy that discuss information exchanges such as segments labeled as  ``First Party Collection/Use", ``Third party sharing/collection", ``Data Retention". Following the steps in Figure~\ref{fig:pipeline}, each segment was split into separate sentences which were annotated by the respective models. 
The results were presented to a human annotator, one of the authors who is an expert on CI. The expert then marked the valid results for each of the privacy statement sentences and CI parameters, and also provided the ground truth. A sentence was marked as a valid flow if it prescribed an information exchange of any kind. Otherwise, by default, all sentences are considered invalid. 

Overall, the extraction phase resulted in a total of  2268 privacy statement sentences, out of which 778 were labeled as valid, containing 3245 CI parameters. On average, a policy contains 18 valid statements, with outliers of 4 and 43 valid statements.

\subsection{HMM and BERT models}

Table~\ref{tbl:HMM_BERT_Supervised} shows the results of training a trigram Hidden Markov Model and a fully-supervised BERT. 
Both models perform relatively poorly for our task, especially when it comes to the ``Sender" parameter. HMM's overall F1 scores are slightly better for detecting other parameters, with the highest F1 score achieved for the TP parameter in both models.

\begin{table}[!ht]
\centering
\begin{tabular}{@{}lp{1.2em}lp{1.2em}lp{1.2em}l@{}}
& \multicolumn{2}{c}{\textbf{Recall}} & \multicolumn{2}{c}{\textbf{Precision}} & \multicolumn{2}{c}{\textbf{F1}} \\
\toprule
\textbf{CI Param.} & {\small HMM} & \small{BERT} & \small{HMM}  & \small{BERT} & \small{HMM}  & \small{BERT} \\
\midrule
Attribute   & 0.65 & 0.59  & 0.59 & 0.43 & 0.62  & 0.50 \\
Receiver   & 0.41  & 0.52  & 0.50  & 0.32    & 0.45 & 0.39 \\
Sender & 0.06   & 0.13   & 0.16     & 0.14    & 0.09 & 0.13 \\
TP     & 0.81   & 0.78   & 0.66     & 0.58    & 0.73 & 0.67     
\end{tabular}
\caption{F1 Scores for fully-supervised HMM and fine-tuned BERT model. The recall and precision values are calculated on word level over the whole test set. }
\label{tbl:HMM_BERT_Supervised}
\end{table}



 





\subsection{DP and SRL}

Table~\ref{tbl:DP_SRL} shows precision and recall for both DP and SRL models\footnote{Specifically we used AllenNLP (0.8.5) with Bert SRL (Bert-base-SRL-2019.06.17)  model.}. Both models have high recall numbers. However, in DP the precision is low, indicating that while DP is able to identify all the relevant instances, it also produces many false positives. SRL performs better, both in terms of precision and recall. The recall numbers are slightly higher compared to DP and the precision is much higher.  We, however, note that SRL did not process 26 statements. They contain verbs that our algorithm  didn't track, some of which are not always associated with information exchange, like ``sell" and ``rent". Figure~\ref{fig:SRL_tag_dist} shows the percentage of SRL arguments that are correctly and incorrectly mapped to the CI parameters. Note that, compared to the dependency tags from DP, semantic arguments from SRL result in more valid mappings to CI parameters. 
\subsubsection{Improved SRL}
Table~\ref{tbl:DP_SRL} shows the results for improved SRL after applying our algorithm incorporating domain-specific heuristics. The precision results have improved across all the parameters, affecting recall only slightly. We note that our F1 metric is calculated on phrase prediction level. 

\begin{table}[!ht]
\centering
\begin{tabular}{p{4em}p{4em}p{2em}p{3.5em}r}
    &    &    {\bf Recall} &  {\bf Precision} &   {\bf F1} \\
{\bf Model} & {\bf CI Param.} & &  & \\
\toprule
DP & Attribute &    0.68 &       0.43 &  0.53 \\
    & Subject &    0.79 &       0.26 &  0.40 \\
    & TP &    0.76 &       0.62 &  0.68 \\
\midrule    
SRL & Attribute &    0.93 &       0.72 &  0.81 \\
    & Receiver &    0.94 &       0.75 &  0.83 \\
    & Sender &    0.95 &       0.64 &  0.76 \\
    & TP &    0.91 &       0.71 &  0.80 \\
\midrule  
CI-SRL & Attribute &    0.91 &       0.77 &  0.83 \\
& Receiver  &    0.88 &       0.79 &  0.84 \\
& Sender    &    0.91 &       0.74 &  0.82 \\
& TP        &    0.90 &       0.84 &  0.87 \\  
\end{tabular}

\caption{F1 Scores for all the models: DP, SRL and Improved SRL (CI-SRL). The recall and precision values for each parameter are calculated by macro averaging over privacy statements.}
\label{tbl:DP_SRL}
\end{table}

\paragraph{Performance across Policies:}

\begin{figure}[!ht]
\centering
		\includegraphics[width=0.59\textwidth]{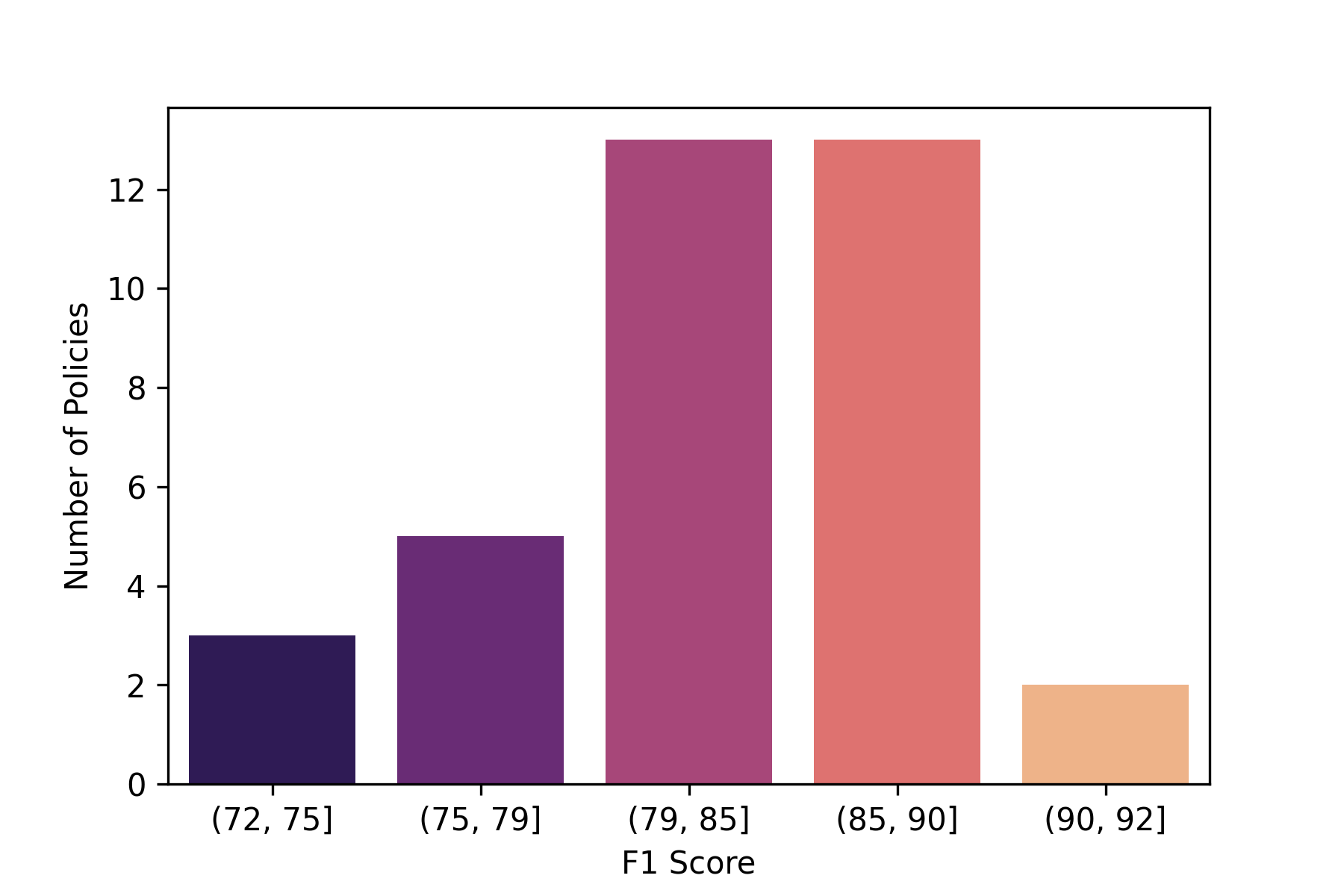}
    \caption{Histogram of F1 scores across privacy policies}
    \label{fig:policy_distribution}
\end{figure}

Figure \ref{fig:policy_distribution} shows F1 score distributions for the annotated policies. The majority of policies (26) have F1 scores in the range of 80-90, which is consistent with the average F1 scores per parameter. A couple of policies perform much better giving a high F1 value of more than 90, whereas some (6) fall in the first group  with F1 scores in the range of 70-79 with 5 out of the 8 policies in the 75-79 F1 range. 

\paragraph{Analysis of Low Precision Statements:}
We analyzed the privacy statements for which our heuristic algorithm achieved low precision scores to better understand the reasons behind the poor performance. The statements where the SRL-based algorithm performed especially poorly involved semantically complex or long connected sentences. Semantically complex statements comprise multiple verb-predicates with related arguments that result in a large number of false positives. For example:
\smallskip\\
\begin{center}
 \noindent\fbox{%
{\scalebox{0.95 }{
    \parbox{1\textwidth}{ 
   { SCEA's consumer services department maintains information {\bf obtained} from consumers who {\bf contact} or {\bf submit} an online complaint so that we may assist these customers with current or future service issues.}
 sensitive personal information {\bf}}.
    }%
}
}
\end{center}
Long connected sentences comprise several phrases. However, due to improper punctuation they appear as a single sentence to our algorithm and as a result generate a large number of false positives. For example, the following statement comprises multiple sentences that are connected with a colon:
\smallskip\\
\begin{center}
 \noindent\fbox{%
{\scalebox{0.95}{
    \parbox{1\textwidth}{ 
    There are two main types of information we collect about users of our online services that include (but are not limited to) the following: Information that identifies you: This is commonly referred to as ``personal information" and includes, for example, information that you provide to us such as your name, home address, age, gender, telephone number, e-mail address, payment information (including your credit card number), and/or photos or video footage of you; and \& Information that relates to you, but on its own does not identify you: Such as information about your Internet connection, the equipment you use to access our online services and information relating to your usage of those services. 
    }
    }%
}
}
\smallskip\\
\end{center}
These cases are not only problematic for an NLP task but also require significant cognitive effort for a human attempting to analyze the privacy implications of the prescribed information flows. Rather than adapting our method to yield better results in these cases, it might be best to use it to detect these complex sentences so that they can be restated more clearly.

\section{Conclusion}

In this paper, we formulate a new CI parameter extraction NLP task for analysis of privacy statements. We adapt several conventional NLP and ML methods (HMM, BERT, DP and SRL) to perform the task and demonstrate that it cannot be solved trivially. In our evaluation of privacy statements from 36 real-world privacy policies, we show that a method combining clues from CI into syntactic DP coupled with type-specific SRL obtains the highest F1 score. We build on this insight to devise an algorithm that incorporates domain-specific knowledge to achieve a much higher precision and recall. The proposed algorithm post-processes ML outputs and increases automation of a tedious task that has so far been performed manually. Further improvements of this task, leveraging domain knowledge for complex scenarios will directly benefit downstream applications ranging from aiding the design and analysis of privacy policies to building systems that meet users' privacy expectations by construction.

\bibliographystyle{coling}
\bibliography{main.bib}

\end{document}